\documentclass[conference, 10pt]{IEEEtran}

\usepackage{cite}
\usepackage{url}
\usepackage{graphicx}
\usepackage{color}
\usepackage{placeins}
\usepackage{float}
\usepackage{tabularx,colortbl}
\usepackage{ifthen}

\usepackage{booktabs}
\usepackage{lipsum}
\usepackage{multirow}
\usepackage{flafter} 

\makeatletter
\newcounter{author}
\renewcommand{\author}[2][]{
   \stepcounter{author}
   \@namedef{author@\theauthor}{#2}
   \@namedef{authorlabel@\theauthor}{#1}
}

\newcounter{address}
\newcommand{\address}[2][]{
   \stepcounter{address}
   \@namedef{address@\theaddress}{#2}
   \@namedef{addresslabel@\theaddress}{#1}
}

\newcommand{\alsep}{and}
\newcommand\figwidth{0.48\textwidth}

\def\newmaketitle{\par%
  \begingroup%
  \normalfont%
  \def\thefootnote{}
  \def\footnotemark{}
  \let\@makefnmark\relax
  \footnotesize
  \footnotesep 0.7\baselineskip
  \normalsize%
  \twocolumn[\thenewmaketitle\@IEEEaftertitletext]%
  \if@IEEEusingpubid
     \enlargethispage{-\@IEEEpubidpullup}%
  \fi
  \endgroup
  \setcounter{footnote}{0}\let\maketitle\relax\let\@maketitle\relax
  \gdef\@thanks{}%
  \let\thanks\relax}

\def\thenewmaketitle{
  \newpage
  \begin{center}%
    \vskip0.2em{\Huge\@IEEEcompsoconly{\sffamily}\@IEEEcompsocconfonly{\normalfont\normalsize\vskip 2\@IEEEnormalsizeunitybaselineskip
   \bfseries\large}\@title\par}\vskip1.0em\par%
    \vspace{1ex}
    \newcounter{c@author}
    \newcounter{c@tmp}
    \ifthenelse{\value{author}=2}{%
      \newcommand{\liand}{ and }}{%
      \newcommand{\liand}{, and }}
    \ifthenelse{\value{address}<2}{%
      \@nameuse{author@1}%
      \stepcounter{c@author}%
      \whiledo{\value{c@author}<\value{author}}{%
        \setcounter{c@tmp}{\value{author}}%
        \addtocounter{c@tmp}{-\value{c@author}}%
        \ifthenelse{\value{c@tmp}=1}{%
          \renewcommand{\alsep}{\liand}}{\renewcommand{\alsep}{, }}%
        \stepcounter{c@author}\alsep \@nameuse{author@\thec@author}}\\%
    }
    {
      \@nameuse{author@1}${}^{(\@nameuse{authorlabel@1})}$%
      \stepcounter{c@author}%
      \whiledo{\value{c@author}<\value{author}}{%
        \setcounter{c@tmp}{\value{author}}%
        \addtocounter{c@tmp}{-\value{c@author}}%
        \ifthenelse{\value{c@tmp}=1}{%
          \renewcommand{\alsep}{\liand}}{\renewcommand{\alsep}{, }}%
        \stepcounter{c@author}\alsep \@nameuse{author@\thec@author}%
        ${}^{(\@nameuse{authorlabel@\thec@author})}$%
      }
    }
    \vspace{0.2ex}

    \ifthenelse{\value{address}>0}{%
      \ifthenelse{\value{address}=1}{
        {\@nameuse{address@1}}
      }
      {
        \newcounter{c@address}

        \begin{center}
        \whiledo{\value{c@address}<\value{address}}
        {
          \refstepcounter{c@address}
            ${}^{(\thec@address)}$\,%
              \label{\@nameuse{addresslabel@\thec@address}}%
              \@nameuse{address@\thec@address}\\ %
        }
        \end{center}
      } 
    }
    {
      \relax
    }
  \end{center}
}

\makeatother
\title{Investigating Map-Based Path Loss Models:\\A Study of Feature Representations in Convolutional Neural Networks}

\author[1, 2]{Ryan G. Dempsey}
\author[1]{Jonathan Ethier}
\author[2]{Halim Yanikomeroglu}

\address[1]{Communications Research Centre (CRC), Ottawa, Ontario, Canada: \{ryan.dempsey, jonathan.ethier\}@ised-isde.gc.ca}
\address[2]{Carleton University, Ottawa, Ontario, Canada: ryandempsey@cmail.carleton.ca, halim@sce.carleton.ca}

\begin{document}

\newmaketitle

\begin{abstract}
Path loss prediction is a beneficial tool for efficient use of the radio frequency spectrum. Building on prior research on high-resolution map-based path loss models, this paper studies convolutional neural network input representations in more detail. We investigate different methods of representing scalar features in convolutional neural networks. Specifically, we compare using frequency and distance as input channels to convolutional layers or as scalar inputs to regression layers. We assess model performance using three different feature configurations and find that representing scalar features as image channels results in the strongest generalization.
\end{abstract}

\section{Introduction}

\IEEEPARstart{O}{ne} of the many emerging enablers supporting the intelligent usage of the radio frequency spectrum in the sixth generation of wireless communications technology (6G) and beyond is machine learning (ML). ML is proving to be an invaluable tool in all parts of communications networks \cite{dl_rl_6g}, including spectrum usage modeling \cite{janaki}, channel coding \cite{coding}, and path loss modeling. Path loss modeling aims to predict point-to-point signal losses from the transmitter (Tx) to the receiver (Rx) of a communications link. It represents a technology-agnostic quantity that can be used to estimate wireless coverage and interference in spectrum planning, licensing, and more. As the number of connected devices grows in future networks and generations, minimizing the error of path loss models becomes a necessity to ensure efficient and seamless use of the available spectrum for all use cases and technologies.

Recent ML-based path loss models have achieved root mean squared error (RMSE) values as low as 6~dB in unseen regions \cite{feature-pl}. These models use relevant engineered features describing a link's obstructions.  These scalar features are derived solely from the direct path and are used as inputs to fully connected networks (FCNs) \cite{Goodfellow-et-al-2016}.

Convolutional neural networks (CNNs) \cite{Goodfellow-et-al-2016} can alternatively be employed to extract complex internal features from obstruction height maps. Using precise surface path profiles as inputs to a CNN consistently provides lower RMSE than the industry standard \cite{P1812} without needing to compute any scalar obstruction features or clutter-derived metrics \cite{pathprofiles}. Though many regions exhibit lower RMSE using this method as compared to the features-only method \cite{feature-pl} by inputting path profiles to a CNN, other cases suffer from overfitting, with higher test RMSE than engineered features in those cases. This indicates the potential for improvement if overfitting can be reduced.

The 2-D CNNs described in \cite{pathprofiles} consist of two main sections: (1) feature extraction layers, which employ convolutional and max pooling layers, and (2) regression layers, which employ fully connected layers to predict path loss using the extracted features. The CNNs use an image-only approach, wherein scalar features are input to the model as image channels \cite{Goodfellow-et-al-2016}, with the entire input being contained in a single 2-D array with an additional channel dimension. This includes the scalar quantity frequency, as a channel filled with a single value for each training and test sample. It also includes a 2-D representation of the distance from the Tx to every point throughout the path profile.

The goal of this paper is to investigate the 2-D CNN-based method proposed in \cite{pathprofiles} by conducting a comprehensive study on the distance and frequency features to determine which representations improve generalization on difficult holdouts. We plan to balance overfitting with practical computational considerations, building models that generalize well with convergence occurring in a reasonable time frame. Specifically, this paper will uncover whether test RMSE is improved by including frequency and distance as inputs to the regression layers, rather than filling entire channels for the feature extraction layers. These features ought to be important, so this study will determine whether including them in the feature extraction layers is a source of overfitting or beneficial to test scores, by comparing with models that omit them from the feature extraction process.

Section~\ref{training} describes the datasets used to train and evaluate CNN-based path loss models. Section~\ref{channels} describes multiple channel structures and feature representations to be considered, which we compare in Section~\ref{models} (including an additional independent test set in a new country). Finally, Section~\ref{conc} summarizes our findings.

\section{Training and Evaluation Data}
\label{training}
\subsection{United Kingdom Data}
The training data involves two data sources: radio measurement drive test data and digital surface model (DSM) data. The drive test data \cite{uk-ofcom} is obtained from the United Kingdom Office of Communications (UK Ofcom) and can be accessed through Ofcom's data repository \cite{open-uk}. Measurements were collected between 2015 and 2018 across six frequency bands (449, 1802, 2695, 3602, and 5850~MHz bands in the UK).

The DSM data is obtained from the UK open government data \cite{uk-dsm-dtm}. It provides surface (trees, buildings, and ground-level variations) heights relative to sea level at a 1~m resolution in various areas in the UK.

We follow the data extraction process outlined in \cite{pathprofiles}. We begin by removing noise-dominated and unreliable measurements \cite[eq. (1)]{pathprofiles}. For each remaining measurement, we then use 2-D nearest neighbor interpolation \cite{splines} to extract a DSM array of shape \((d\times W)\) around the direct path from the DSM data (where \(d\) is the link distance in metres, rounded down, and \(W = 61\ \mathrm{m}\), to be consistent with \cite{pathprofiles}). The Tx and Rx are in the center of the array, resulting in \((W-1)/2\ [\mathrm{m}]\) of obstruction heights in the array around each side of the direct path. After accounting for Earth curvature with a radius of 6365~km as described in \cite{pathprofiles}, the array is finally re-sampled to shape \((256\times W)\) using bilinear interpolation \cite{bilinear}. This array is used as one channel of an input image of shape \((C\times256\times W)\), where \(C\) denotes the number of channels. The second channel is generated by computing the height of the direct path between the antennas, and inputting this in the center row of the array, with all other rows being 0. These two channels serve as the baseline structure with the remaining features being compared in this paper.

\subsection{Additional Test Data}
As a more rigorous evaluation of generalization, we use proprietary measurement data provided by NetScout \cite{NetScout}. This dataset features measurements from varied areas across Canada, including dense urban, suburban, and rural areas.

Ten distinct regions are used, with frequencies between 700 and 3700~MHz. We obtain the Canadian path profiles using the same method as in the UK, using the High Resolution Digital Elevation Model (HRDEM) \cite{HRDEM} for DSM data. After performing the aforementioned processing steps and only using samples with 1~m resolution HRDEM, nearly 125\,000 measurements remain for testing.

\section{Channel Configurations}
\label{channels}
The CNNs proposed in \cite{pathprofiles} used an image-only approach, where each input feature formed a channel of constant size. The model used four channels (DSM, height of direct path from Tx to Rx, frequency, and 2-D Euclidean distance). The normalization methods can be seen in \cite{pathprofiles}. Euclidean distance from the Tx is computed at every pixel in the image to provide spatial information \cite{coordconv} relative to the Tx, while also providing the non-re-sampled link distance. This may lead to overfitting due to the high number of features \cite{overfitting}.

In this paper, we determine whether having a simpler yet more informative distance feature reduces overfitting. As an alternative representation of distance, we compute 3-D Euclidean distance from Tx to Rx as a single scalar, and compare its inclusion as either a filled channel of constant value or a scalar feature input to the FCN. To have a consistent representation of scalar information, we also include frequency as a scalar with the distance scalar. This will indicate whether scalar features should be included as channels or scalars for path loss modeling.

\begin{figure*}
    \centering
    \includegraphics[width=0.75\textwidth]{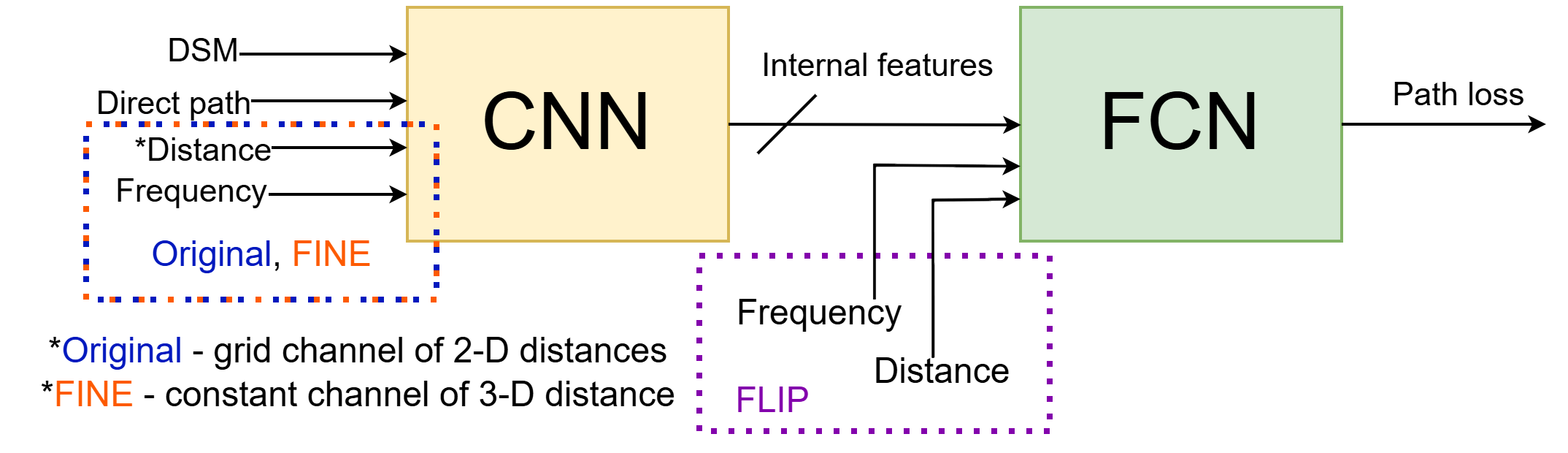}
    \caption{Convolutional neural network topology. *Original and FINE both use a filled frequency channel, with FINE replacing the original 2-D distance grid with a filled channel containing 3-D link distance. FLIP represents both frequency and distance as scalar inputs to the FCN.}
    \label{topology-fig}
\end{figure*}

Table~\ref{channels-table} and Fig.~\ref{topology-fig} show the channel configurations we compare in this paper. We compare the original configuration to including both frequency and 3-D distance as filled channels of constant value as well as including both frequency and 3-D distance as scalar features. Alongside the original configuration, we consider the following two configurations:
\begin{itemize}
    \item Feature Integration for Non-linear Extraction (FINE).\\This model uses both 3-D link distance and frequency as single scalar features filling their respective channels.
    \item Feature Layer Integration in Perceptron (FLIP).\\This model uses both 3-D link distance and frequency as single scalar features integrated in the FCN (i.e. perceptron).
\end{itemize}

\begin{table}[!t]
\centering
\caption{Channel Configuration for Each Model}
\label{channels-table}
\begin{tabular}{lccc}
\toprule
Feature & Original & FINE & FLIP \\
\midrule
Distance & 2-D Grid channel & 3-D Filled channel & 3-D Scalar \\
Frequency & Filled channel & Filled channel & Scalar \\
\midrule
\# of channels & 4 & 4 & 2 \\
\# of scalars & 0 & 0 & 2 \\
\bottomrule
\end{tabular}
\end{table}

\section{Model Performance}
\label{models}
\subsection{Cross-Validation in the United Kingdom}
We use the cross-validation approach outlined in \cite{pathprofiles}, where each city acts as an independent holdout, with each of the remaining five cities individually geographically split between training and validation (80\% and 20\% respectively). We perform 10 optimization runs for each holdout. All runs use adaptive moment estimation (Adam) with a learning rate of 0.0001 to optimize mean squared error, with a batch size of 256, trained for 200 epochs. In a given run, the model checkpoint with the lowest validation loss is used to evaluate test RMSE.

The cross-validation results are summarized in Table~\ref{results-combined}. Simplifying the distance feature to a single 3-D distance has positive results on Merthyr Tydfil, Southampton, and Stevenage holdouts, as shown in the FINE mean column of Table~\ref{results-combined}. However, this configuration is to the major detriment of the other three holdouts, with far higher mean RMSE in those cases. The mean and standard deviation (SD) of RMSE averaged across all cities are both higher, indicating that the new configuration hinders more than it helps.

\begin{table}[!t]
\centering
\caption{Cross-Validation Test RMSE (\textnormal{d}B) (10 Runs)}
\label{results-combined}
\begin{tabular}{lcccccc}
\toprule
        & \multicolumn{2}{c}{Original \cite{pathprofiles}} & \multicolumn{2}{c}{FINE} & \multicolumn{2}{c}{FLIP} \\
       Holdout & Mean & SD & Mean & SD & Mean & SD\\
\midrule
        Boston & \textbf{7.18} & 0.27 &          7.76 & 0.79 & 22.34 & 1.46\\
        London & \textbf{7.75} & 0.47 &          8.22 & 0.60 & 12.56 & 0.90\\
Merthyr Tydfil &          8.07 & 0.39 & \textbf{7.75} & 0.28 & 12.67 & 0.56\\
    Nottingham & \textbf{6.99} & 0.23 & 7.10 & 0.20 & 12.09 & 0.32\\
   Southampton &          6.35 & 0.42 & \textbf{6.31} & 0.36 & 10.28 & 0.56\\
     Stevenage &          7.73 & 0.21 & \textbf{7.62} & 0.26 & 12.52 & 0.57\\
\midrule
          Mean & \textbf{7.35} & 0.33 &          7.46 & 0.42 & 13.74 & 0.73\\
\bottomrule
\end{tabular}
\end{table}

Inputting the scalars after the convolutional layers greatly increases test RMSE, as shown in the FLIP column of Table~\ref{results-combined}. Although the FLIP training RMSE achieves similar results to the other two configurations (\textless~5~dB), the validation and test RMSE are consistently much higher than the other two models (\textgreater~10~dB). This result indicates overfitting.

Since frequency affects the electromagnetic interactions with clutter, it is possible that frequency is optimally present before the convolutional layers, in order to influence the extraction of relevant, generalizable GIS features from the DSM data. Similarly, the distance feature may need to be present during feature extraction to provide the scale of obstructions, since the inputs are re-sampled to a constant size. This may explain FLIP's high test RMSE, having these two features in the regression layers rather than the feature extraction layers. We conclude that both scalar features, frequency and distance, should be used as channels (as in the original and FINE models) in CNN-based path loss models rather than scalar inputs to the FCN (as in the FLIP model), to minimize test RMSE.

\subsection{No-Holdout Models and Blind Testing in Canada}
\label{tests}
The results presented in Table~\ref{results-combined} demonstrate strong generalization in the original and FINE models, compared to the FLIP model. This indicates that representing frequency and distance as channels before the feature extraction layers is helpful in training generalized models. This section tests that idea more thoroughly using independent measurement data in Canada.

As a final test, we re-train all three models using all six UK cities in training and validation (hereafter referred to as ``no-holdout'' models). We again perform 10 optimization runs for each model, resulting in 10 different trained models per configuration. We create an ensemble of the 10 no-holdout models using simple averaging \cite{ensembling} (taking the arithmetic mean of the 10 outputs for a given input). Table~\ref{netscout-test} summarizes the test results. In addition to computing RMSEs for each city, we also group all measurements into dense urban, suburban, and rural, and break down the results by these categories.

\begin{table}[!t]
\centering
\caption{10-Run No-Holdout Ensemble Test RMSE (\textnormal{d}B)}
\label{netscout-test}
\begin{tabular}{lccc}
\toprule
       Region & Original \cite{pathprofiles} & FINE & FLIP \\
\midrule
       Halifax & \textbf{7.82} &          7.87 & 12.09\\
   Laurentides &          5.07 & \textbf{4.95} &  8.64\\
   Mississauga &          7.54 &          7.51 &\textbf{7.26}\\
       Moncton & \textbf{4.34} &          4.35 & 5.46\\
 New Brunswick &          5.26 & \textbf{4.83} & 14.28\\
  Peterborough &          6.42 & \textbf{6.23} & 10.86\\
    Sherbrooke &          8.68 & \textbf{8.30} & 11.14\\
       Toronto &          8.01 & \textbf{7.91} &  9.35\\
        Whitby &          7.97 &          7.32 &\textbf{6.77}\\
      Winnipeg & \textbf{9.02} &          9.14 &\textbf{8.54}\\
\midrule
   Dense urban & \textbf{8.34} &          8.39 & 10.50\\
         Rural &          5.84 & \textbf{5.59} & 11.80\\
      Suburban &          7.82 & \textbf{7.52} &  8.68\\
\midrule
\midrule
       Overall &          7.42 & \textbf{7.25} & 10.26\\
\bottomrule
\end{tabular}
\end{table}

\begin{figure}[!t]
\centering
\includegraphics[width=\figwidth]{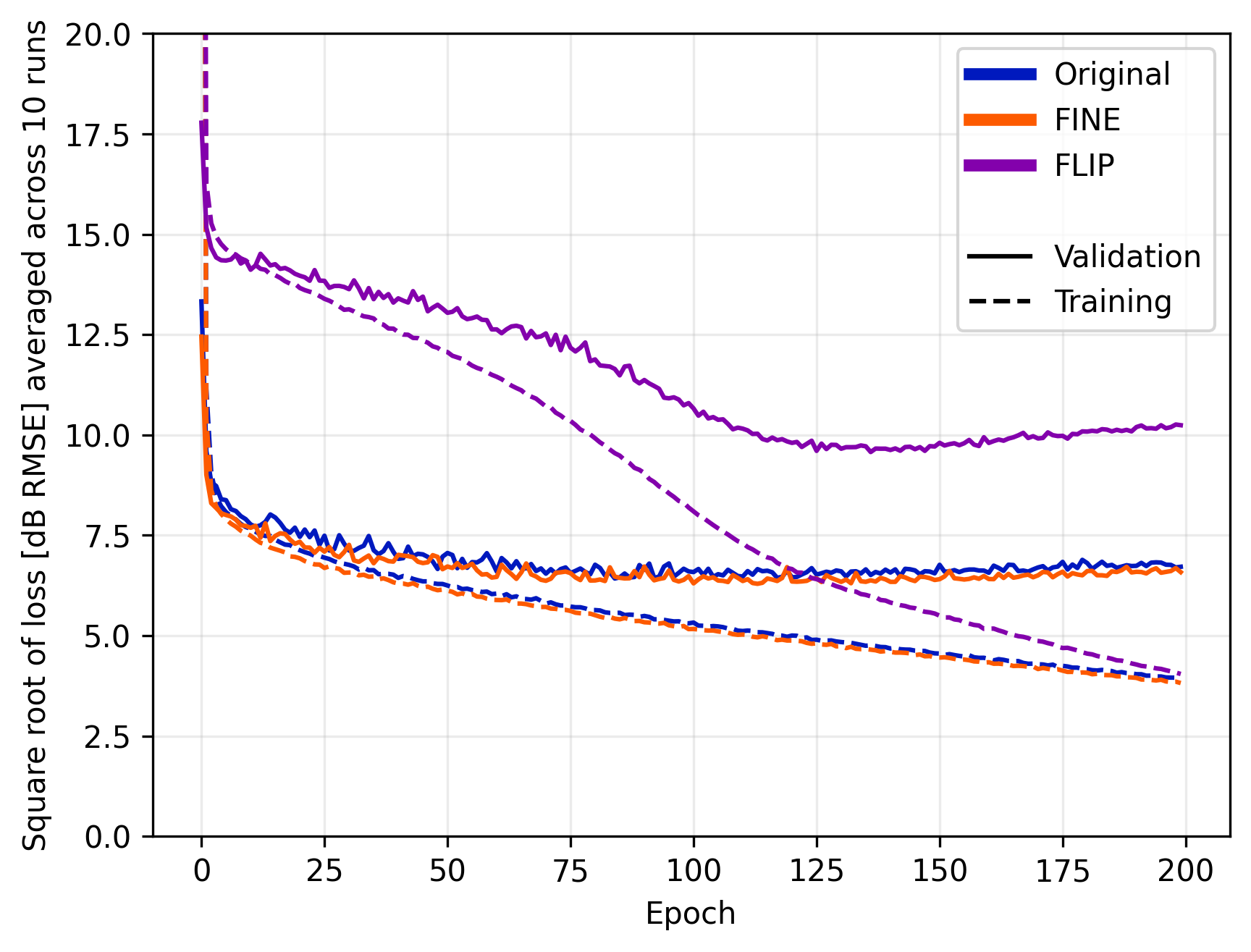}
\caption{Average loss curve across all 10 no-holdout models.}
\label{losscurve-netscout}
\end{figure}

Fig.~\ref{losscurve-netscout} shows the average loss curve for the no-holdout models, by taking the mean RMSE across all 10 runs in a given epoch. Evidently, FINE no longer overfits when provided all six cities in both training and validation - both validation and training RMSE are lower than that of the original. It is possible that with the increased sample diversity (six cities instead of five), simpler features can provide stronger performance and generalization, as is the case with the single 3-D distance feature seen in FINE (7.25~dB RMSE) compared to the dense 2-D distance grid seen in the original (7.42~dB RMSE). All three models appear to have benefited (lower validation and test RMSE) from the increased training set and ensemble modeling. This includes the FLIP model, which is discussed in further detail below.

\subsection{Investigating FLIP Performance}
On the Canadian data, FLIP demonstrates a turnaround in performance, exhibiting the lowest ensemble RMSE on three of ten holdouts. Similarly to the other two models, FLIP's RMSE decreased from Table~\ref{results-combined} to Table~\ref{netscout-test}, likely for similar reasons. To investigate the causes of FLIP's successes and failures, we conduct an exhaustive search of features for linear regression where the target is FLIP RMSE per city. We compute the coefficient of determination R$^2$ for all combinations of the following three features: proportion of links having line of sight ($p_{LOS}$), mean total obstruction depth [m] ($\mu_o$), and SD of link distance~[m] ($\sigma_d$). After testing several variables, these three features consistently contributed to high R$^2$ among the various combinations tried, with diminishing returns using more than three. We normalize all features to [0, 1] before each regression. We summarize the results in Table~\ref{regression-table}.

\begin{table}
    \centering
    \caption{Feature(s) Correlated with FLIP RMSE by City}
    \label{regression-table}
    \begin{tabular}{rrl}
    \toprule
        Features Used in Linear Regression & R$^2$ & Coefficient(s)\\
    \midrule
        $\sigma_d$ & 0.02 & $1.30$ \\
        $\mu_o$ & 0.05 & $1.90$ \\
        $p_{LOS}$ & 0.47 & $6.68$ \\
        $\mu_o + \sigma_d$ & 0.05 & $2.48, -0.78$ \\
        $p_{LOS} + \sigma_d$ & 0.47 & $6.77, -0.32$ \\
        $p_{LOS} + \mu_o$ & 0.56 & $7.03, 2.60$ \\
        $p_{LOS} + \mu_o + \sigma_d$ & 0.90 & $11.18, 10.87, -10.49$ \\
    \bottomrule
    \end{tabular}
\end{table}

Comparing the signs of the coefficients in Table~\ref{regression-table}, a higher proportion of line-of-sight links and a higher obstruction depth in a given city both increase RMSE. Conversely, a higher SD of link distance generally decreases RMSE. Though one row contains a positive $\sigma_d$ coefficient, the corresponding $R^2$ is very low, discrediting the result. One possible explanation for the coefficients is that the FLIP model achieves low error specifically on non-line-of-sight yet low clutter density links. There are very few line-of-sight links in the UK training set (1.5\%), so all three models including FLIP likely struggle with purely line-of-sight links. However, the presence of less clutter may reduce the errors imposed by FLIP's GIS feature extraction described above. Therefore, a small, yet non-zero clutter density may minimize FLIP's error.

Larger SDs of link distance decrease FLIP error. After examining the data, we note that the longest links produce the lowest FLIP error. We speculate that since the longest links are close to the maximum path loss of the system, there is a much tighter range of potential values, leading to more frequent ``lucky guesses'', and a lower RMSE.

These coefficient interpretations are speculative. There is undoubtedly more work to be done in analyzing the strengths and weaknesses of all three models to minimize RMSE.

\section{Concluding Remarks}
\label{conc}
This paper provided a comprehensive analysis of best practices in representing scalar features in deep learning-based path loss models using path profiles. We found that representing simple scalar features using filled image channels can achieve 7.25~dB RMSE in a continentally independent test set of 125\,000 measurements. We also examined one model's strengths and weaknesses after inconsistent RMSE in independent testing. Future studies will investigate best practices in including supplemental features to further improve the model fit without sacrificing generalization. This may include terrain heights, Fresnel zone information, and clutter types (tree types, building types, etc.). We also plan to further examine the current models' predictive strengths and weaknesses.


\bibliographystyle{IEEEtran}
\bibliography{references} 

\end{document}